\documentclass{article}
\usepackage{spconf,amsmath,graphicx}
\usepackage[skip=0.05\baselineskip]{caption}
\usepackage{color}
\usepackage{multirow}
\usepackage{multicol}
\usepackage{enumitem}
\setenumerate[1]{itemsep=0pt,partopsep=0pt,parsep=\parskip,topsep=5pt}
\setitemize[1]{itemsep=0pt,partopsep=0pt,parsep=\parskip,topsep=5pt}
\setdescription{itemsep=0pt,partopsep=0pt,parsep=\parskip,topsep=5pt}

\title{SM3D: SIMULTANEOUS Monocular MAPPING AND 3D DETECTION}
\name{Runfa Li, Truong Nguyen\thanks{This project is supported in part by the SRIP program at ECE Dept., UCSD.}}
\address{Department of Electrical and Computer Engineering, University of California, San Diego}
\begin{document}
%\ninept
%
\maketitle
\begin{abstract}
\label{sec:abs}

Mapping and 3D detection are two major issues in vision-based robotics, and self-driving. While previous works only focus on each task separately, we present an innovative and efficient multi-task deep learning framework (SM3D) for Simultaneous Mapping and 3D Detection by bridging the gap with robust depth estimation and “Pseudo-Lidar” point cloud for the first time. The Mapping module takes consecutive monocular frames to generate depth and pose estimation. In 3D Detection module, the depth estimation is projected into 3D space to generate “Pseudo-Lidar” point cloud, where Lidar-based 3D detector can be leveraged on point cloud for vehicular 3D detection and localization. By end-to-end training of both modules, the proposed mapping and 3D detection method outperforms the state-of-the-art baseline by 10.0\% and 13.2\% in accuracy, respectively. While achieving better accuracy, our monocular multi-task SM3D is more than 2 times faster than pure stereo 3D detector, and 18.3\% faster than using two modules separately. 
\end{abstract}
\section{Introduction}
\label{sec:intro}

Traditional mapping and visual odometry strategies are mostly based on SLAM \cite{SLAM,orb_slam} (Simultaneous Localization and Mapping) algorithm, which is well-known for simultaneously perceiving surrounding environments and keeping track of the ego motion.  However, traditional SLAM requires ubiquitous sensors, expensive depth cameras which are not only high-cost but also computationally expensive. SFM (Structure from motion) is a good alternative to SLAM, using only consecutive image snippets. Using an elegant self-supervised learning style, SFM Learner\cite{SFM_learner} jointly trains the depth model and pose model with photo-consistency loss between target and warped images. Monodepth2\cite{monodepth2} introduces SSIM\cite{SSIM} in SFM to enforce photometric consistency, filtering out lighting changes and imaging noises. Similarly, \cite{perceptual_loss} adds perceptual loss by calculating the pixel-wise CNN features.  Since learning of mapping by scene flow assumes that the background is completely rigid, many supervised methods design segmentation mask to eliminate dynamic objects \cite{segmask_1, segmask_2}, while unsupervised methods rely on “soft mask” by adding optical self-supervision in the training process. DF-Net\cite{DF-Net} uses a pretrained optical flow network to segment non-rigid objects, while GeoNet\cite{geonet} employs a subsequent module to compensate for the final predicted motion estimation. 
\iffalse
However, to the best of our knowledge, all previous works ignore a potential constraint existing in the input monocular snippet, we introduce a novel pose-consistency loss.
\fi

Existing 3D detection algorithms mostly use 2D-3D prototypes \cite{Mono3D,Deep3DBox,shift_rcnn}, which are based on 2D object detection, where different geometric constraints are imposed to project 2D proposals to 3D.  Although these approaches give reasonable 3D proposals, they lack in producing accurate 3D location. Alternatively, Lidar-based methods \cite{pointrcnn,fast_pointrcnn,pointpillar} are far more accurate than state-of-the-art 2D-3D prototypes. Similar to mapping, latest detection works try to eliminate expensive Lidar, Radar, and depth camera, by using only images. The Lidar-based detection prototype could be directly applied to images by depth estimation, which saves the cost on device and maintains high accuracy. Based on the experimental results, we inherit the Lidar-based style and use monocular-based depth for end-to-end training which further improve the performances over state-of-the-art monocular-based models while achieve much higher efficiency compared to stereo-based models \cite{3D_proposal,PL-MONO}.

Since the previous works estimate the ego-map and 3D detection separately, they ignore the potential to efficiently integrate a multi-task model. Our key contributions can be briefly summarized as follows:
\begin{itemize}[leftmargin=0.5cm]
\setlength{\itemsep}{0pt}
\setlength{\parsep}{0pt}
\setlength{\parskip}{0pt}
\item We propose a multi-task framework that only takes monocular inputs to estimate ego-map and 3D detection simultaneously by bridging the gap with robust depth estimation and Pseudo-Lidar generation. 
\end{itemize}

\begin{itemize}[leftmargin=0.5cm]
\setlength{\itemsep}{0pt}
\setlength{\parsep}{0pt}
\setlength{\parskip}{0pt}
\item
We derive and impose a novel pose-consistency loss in the end-to-end self-supervised training for Mapping Module while keeping the efficiency in testing, which significantly boosts the mapping performance.
\end{itemize}

\begin{itemize}[leftmargin=0.5cm]
\setlength{\itemsep}{0pt}
\setlength{\parsep}{0pt}
\setlength{\parskip}{0pt}
\item
To the best of our knowledge, this work is the first work that uses monocular input to successfully train a 3D detector end-to-end.  The resulting 3D Detection module outperforms the state-of-the-art 3D detectors with monocular inputs.
\end{itemize}

\begin{figure*}[t]
    \centering
    \includegraphics[width=\linewidth]{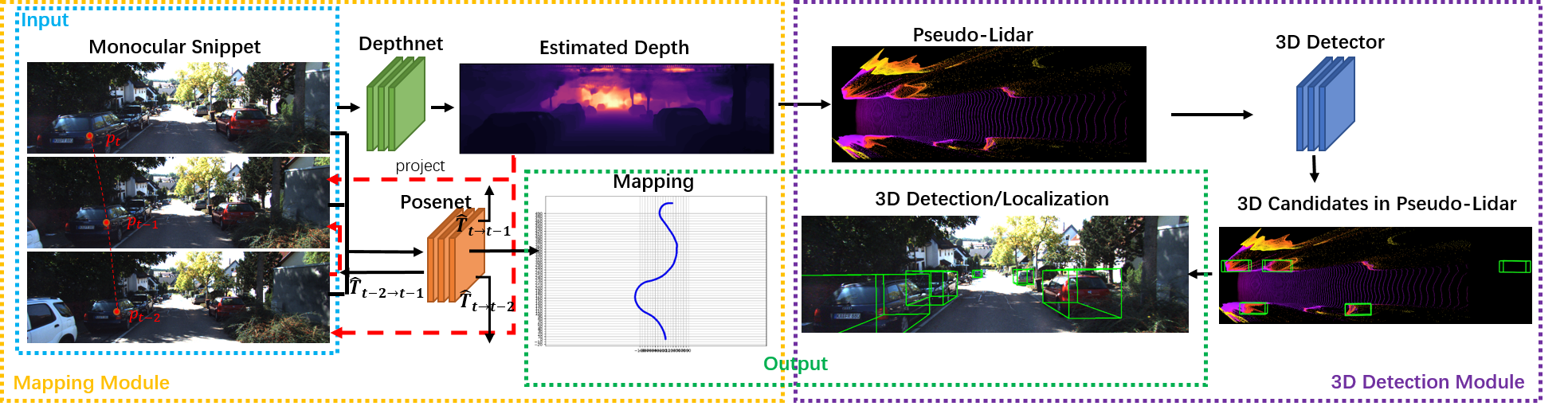}
    \caption{Overview of our SM3D. Mapping Module: Jointly learning and estimating depth and pose. 3D Detection Module: Jointly learning and estimating depth and 3D detection. Input: Monocular snippet. Output: Mapping/3D Detection \& Location.} 
    \label{fig:my_label}
    \vspace{-5mm}
\end{figure*}

\iffalse
\begin{figure*}
\begin{minipage}[h]{0.5\textwidth}
    \centerline{\includegraphics[width=1.0\textwidth]{images/methodology.png}}
    %\vspace*{1.5mm}
    \caption{Illustration of the image warping process, from \cite{ren2017unsupervised}}
    \label{fig:warping}
\end{minipage}
\end{figure*}
\fi

\section{SM3D Framework}
\label{sec:format}
In this section, we present the proposed SM3D network with Mapping and 3D detection modules (See Fig. 1). The input is real-time monocular snippet of consecutive frames, the outputs are real-time ego-map to the current frame and 3D object detection/localization of the current frame.

\subsection{Mapping Module}
We design our Mapping Module following the success of the concurrent SFM works, which use a similar prototype based on the baseline SFM Learner\cite{SFM_learner}. This module takes a snippet of consecutive monocular frames, jointly trains a depth and pose network in a self-supervised manner.

We use the following transformation to project a pixel from the target view $p_{t}$ to the source view $p_{s}$:  
\begin{equation}
    p_{s} = K\hat{T}_{t\to s}\hat{D}_t(p_t)K^{-1}p_{t}
 \end{equation}
where $K$ is the camera intrinsics, $\hat{T}_{t\to s}$ is the pose estimation from target view to source views, $\hat{D}_t(p_t)$ is the estimated depth of pixel $p_t$ from depth network. Bilinear interpolation is then used to populate discrete value for coordinates of the projected pixels. 
\iffalse
The transformation enables the projection from target view to source view, the inverse transformation also enables to reconstruct the target view from the source view, now that the ground truth source views originally exist in frame sequences, 
\fi
The reconstructed target view can be produced with the inverse transformation, then the photometric consistency loss can be defined as: 

\begin{equation}
    \mathcal{L}_{vs}=\sum_{s}\sum_{p}|I_t(p)-\hat{I}_s(p)|
\end{equation}

\noindent where p is the index of the pixel coordinates, \(I_t\) is the target view and \(I_s\) are the source view. \(\hat{I}_s\) is the reconstructed target view inverse-transformed from the source views.  We improve the SFM model in two aspects.

First, to the best of our knowledge, previous SFM works use disparity network. However, if the disparity network is trained to estimate depth, its intrinsic error will be exacerbated for far-away objects \cite{PL++}. Under such concern, our SM3D initializes with a depth network, which leads to better performance.

Second, to improve the long-term robustness of pose estimation, many previous works increase the length of the input sequence which significantly increases the computation costs in both the training and testing process.  Other works randomly sample the neighbor frames in a longer time distance while keeping the length of sequence. However these approaches ignore the inner connection of a snippet in a given length.   To take advantage of photometric consistency while still saving the computation cost, we derive a new constraint on the total loss. In the simplest case, when the snippet length is 3, where frame $T$ is the target view, and frames $T-1$, $T-2$ as the source views, the proposed skip-time pose consistency constraint can be described as:

\begin{equation}
    \hat{T}_{t-1\to t} \hat{T}_{t-2\to t-1} = \hat{T}_{t-2\to t}
    \label{loss2}
\end{equation}

\noindent Here $\hat{T}_{m\to n}$ is the estimated pose from frame m to n. When using longer snippet, it will be extended between all frames, which maximally utilize the pose consistency while saving the computation cost in testing by only imposing in training process.

\subsection{3D Detection Module }
3D Detection Module is designed in a Pseudo-Lidar approach that jointly trains depthnet and 3D detector, in an end-to-end approach.  We generate Pseudo-Lidar from depth estimation by projecting each 2D pixel to 3D space. Given 2D coordinate of each pixel \((u,v)\) in the depth map, the projection process can be derived as:
\begin{gather}
    z = D(u,v), \; x = \frac{(u-c_U)\times z}{f_U}, \; y = \frac{(v-c_V)\times z}{f_V}
    \label{eq:3}
\end{gather}
where $z, x, y$ are the depth, width and height of the corresponding projected point in 3D coordinate. \((c_U,c_V)\) is the pixel location w.r.t the camera center and \(f_V\) is the vertical focal length of the camera.  

Since the Pseudo-Lidar generation process is clearly differentiable w.r.t $z$ in Eq. (4), the end-to-end training is applicable by back-propagating the loss from 3D detector all the way back to the monocular depthnet. Similar to \cite{e2e_PL}, the jointly learning loss is defined as:  

\begin{equation}
    \mathcal{L}=\lambda_{det}\mathcal{L}_{det} + \lambda_{depth}\mathcal{L}_{depth}
\end{equation}

\noindent where $\mathcal{L}_{det}$ and $\mathcal{L}_{depth}$ are the loss of 3D detection and depth estimation, $\lambda_{det}$ and $\lambda_{depth}$ are corresponding weights. $\mathcal{L}_{det}$ is a combination of the classification loss for candidate category and the regression loss for the bounding box location. $\mathcal{L}_{depth}$ is the L1 loss of the estimated depth and the ground truth.

\subsection{SM3D Model}
Finally, we combine the trained Mapping Module and 3D Detection Module together to build our SM3D network. For the utility purpose, since the posenet from Mapping Module can be used independently of the jointly trained depthnet, we use the depthnet jointly trained with 3D detector for SM3D.

\section{Experimental results}
\label{sec:pagestyle}
\begin{table}[]
\caption{Absolute Trajectory Error (ATE) on the KITTI odometry test split averaged over all 3-frame snippets, all models are trained with snippet length 3 on our subset of KITTI odometry training split.}
\begin{tabular}{ccc}
\hline
Method               & Seq.  09               & Seq. 10                \\ \hline
SFM Learner          & 0.0100 $\pm$ 0.0063          & 0.0085 $\pm$ 0.0073          \\
\textbf{SM3D (ours)} & \textbf{0.0090 $\pm$ 0.0052} & \textbf{0.0084 $\pm$ 0.0067} \\ \hline
\end{tabular}
\vspace{-4mm}
\end{table}

\begin{figure}
\begin{minipage}{1.0\columnwidth}
\begin{minipage}{0.49\columnwidth}
  \centerline{\includegraphics[width=1.0\linewidth]{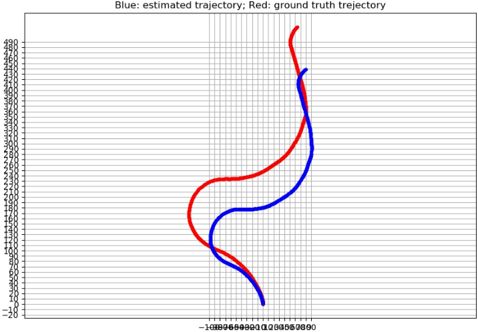}}
  \centerline{(a) SuperPoint}
\end{minipage}
\hfill
\begin{minipage}{0.49\columnwidth} 
  \centerline{\includegraphics[width=1.0\linewidth]{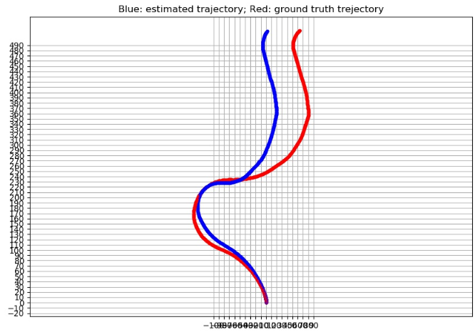}}
  \centerline{(b) Libviso2}
\end{minipage}
\quad
\end{minipage}

\begin{minipage}{1.0\columnwidth}
\begin{minipage}{0.49\columnwidth}
  \centerline{\includegraphics[width=1.0\linewidth]{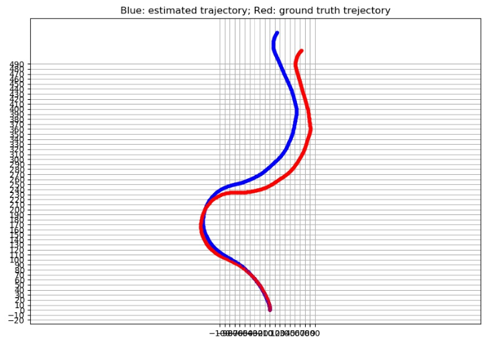}}
  \centerline{(c) SFM Learner}
\end{minipage}
\hfill
\begin{minipage}{0.49\columnwidth} 
  \centerline{\includegraphics[width=1.0\linewidth]{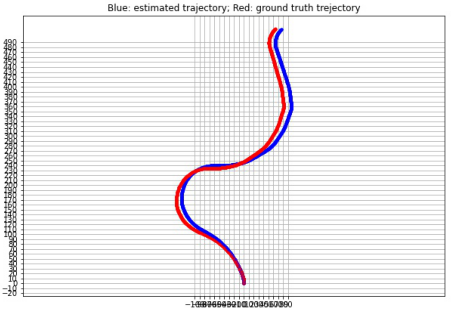}}
  \centerline{(d) SM3D (Ours)}
\end{minipage}
\quad
\end{minipage}

\caption{Mapping results on the first 700 frames of sequence 09, KITTI test split. The ground truth trajectory is in \textcolor{red}{red}, the estimated trajectory is in \textcolor{blue}{blue}. The estimated trajectory is recovered from the estimated pose.}
\label{mapping}
\vspace{-5mm}
\end{figure}

\subsection{Setup}
We use KITTI dataset to evaluate the algorithm performance. On the Mapping Module, for simplicity, we use a snippet length of 3, and train all models on a subset of KITTI odometry training split. Similar to previous works, we evaluate on sequence 09 and 10 of KITTI odometry test split. For the 3D Detection Module, we train and evaluate with KITTI 3D detection benchmark, with 3712, 3769, 7518 images for training, validation and testing, respectively. We use two Nvidia GTX 1080Ti GPU for training. For end-to-end training of the Mapping Module, we initialize the depthnet from a pretrained BTS network \cite{BTS}, while the posenet is jointly trained from scratch using same network as SFM Learner. The image size is $128\times 416$, the learning rate is set to $2\times10^{-4}$ with batchsize 4, the parameters $\alpha$ and $\beta$ of Adam optimizer are set to 0.9 and 0.999. For end-to-end training of the 3D Detection Module, we first initialize from pretrained BTS and choose PointRCNN as our 3D detector. Similar to \cite{e2e_PL}, we first fix depthnet to train the 3D detector from the scratch. Finally, we jointly train the detector and fine-tune the pretrained BTS depth network. The depth ground truth is projected from Lidar data in KITTI. The image size is $352 \times 1216$,  For testing, we use a single Nvidia GTX 1080 Ti GPU.

\subsection{Qualitative Results}
\textbf{Mapping Module.}
Based on the Absolute Trajectory Error (ATE) reported in Table 1, the proposed SM3D network is 10.0\% and 1.2\% better than the baseline SFM learner on sequence 09 and 10 respectively. Recovered from the estimated pose, we visualize the mapping trajectory as shown in Fig 2. SuperPoint\cite{superpoint}, Libviso2\cite{libviso2}, and SFM Learner\cite{SFM_learner} are chosen for comparison, where all models are trained on our data split. As observed, the proposed SM3D network has the closest trajectory to the ground truth, which outperforms all methods above. Such results further validate our design of the skip-time photometric consistency constraint and the utility of depth rather than disparity.

\begin{table*}[t]
\begin{center}
\caption{Qualitative detection results comparison on KITTI val set. We report the average precision (in \%) of car category on bird’s eye view and 3D object detection as AP$_{BEV}$ and AP$_{3D}$. Top two rows are state-of-the-art "2D-3D" methods, middle three rows are concurrent 3D methods, bottom two rows in \textcolor{blue}{blue} are state-of-the-art concurrent Pseudo-Lidar based 3D methods. The proposed SM3D (in \textcolor{red}{red}) outperforms all monocular methods at IoU=0.7. AP at lower IoU threshold 0.5 only reflects the 3D box proposal/candidates, while the higher IoU threshold at 0.7 reflects the precision of the box coordinates, thus the higher AP at IoU=0.7 validates that our SM3D is better for 3D localization, not just for 3D proposal.}  
\begin{tabular}{|c|c|c|c|c|c|c|c|}
\hline
\multirow{2}{*}{Method}                                        & \multirow{2}{*}{Input} & \multicolumn{3}{c|}{AP$_{BEV}$/AP$_{3D}$ (in \%),  IoU = 0.5} & \multicolumn{3}{c|}{AP$_{BEV}$/AP$_{3D}$ (in \%), IoU = 0.7}           \\ \cline{3-8} 
                                                               &                        & Easy            & Moderate        & Hard            & Easy               & Moderate           & Hard               \\ \hline
Mono3D\cite{Mono3D}                                                         & Monocular              & 30.5/25.2       & 22.4/18.2       & 19.2/15.5       & 5.2/2.5            & 5.2/2.3            & 4.1/2.3            \\
Deep3DBox\cite{Deep3DBox}                                                      & Monocular              & 30.0/27.0       & 23.8/20.6       & 18.8/15.9       & 10.0/5.6           & 7.7/4.1            & 5.3/3.8            \\ \hline
MLF-MONO\cite{MLF-MONO}                                                       & Monocular              & 55.0/47.9       & 36.7/29.5       & 31.3/26.4       & 22.0/10.5          & 13.6/5.7           & 11.6/5.4           \\
ROI-10D\cite{ROI-10D}                                                        & Monocular              & 46.9/37.6       & 34.1/25.1       & 30.5/21.8       & 14.5/9.6           & 9.9/6.6            & 8.7/6.3            \\
MonoGRNet\cite{MonoGRnet}                                                      & Monocular              & -/50.5          & -/37.0          & -/30.8          & -/13.9             & -/10.2             & -/7.6              \\ \hline
{\color{blue} PL-MONO-FP\cite{PL-MONO}}                                                     & Monocular              & 70.8/66.3       & 49.4/42.3       & 42.7/38.5       & 40.6/28.2          & 26.3/18.5          & 22.9/16.4          \\
{\color{blue} Mono3DPLiDAR\cite{Mono3DPLiDAR}}                                                   & Monocular              & 72.1/68.4       & 53.1/48.3       & 44.6/43.0       & 41.9/31.5          & 28.3/21.0          & 24.5/17.5          \\ \hline
\begin{tabular}[c]{@{}c@{}}Naive SM3D (Ours)\end{tabular}    & Monocular              & 70.7/45.1       & 52.5/31.2       & 47.3/22.6       & 38.0/5.4           & 29.6/4.6           & 26.2/4.5           \\
{\color{red} \textbf{SM3D (Ours)}} & Monocular              & 68.1/65.9       & 49.1/47.2       & 41.3/40.0       & {\color{red}\textbf{45.8/33.1}} & {\color{red}\textbf{32.8/23.9}} & {\color{red}\textbf{28.0/20.4}} \\ \hline
PL-STEREO-FP\cite{PL-MONO}                                                  & Stereo                 & 89.8/89.5       & 77.6/75.5       & 68.2/66.3       & 72.8/59.4          & 51.8/39.8          & 44.0/33.5          \\ \hline
\end{tabular}
\end{center}
\textbf{\vspace{-7mm}}
\end{table*}

\begin{figure*}
\begin{minipage}{0.33\textwidth}
  \centerline{\includegraphics[width=1.0\textwidth]{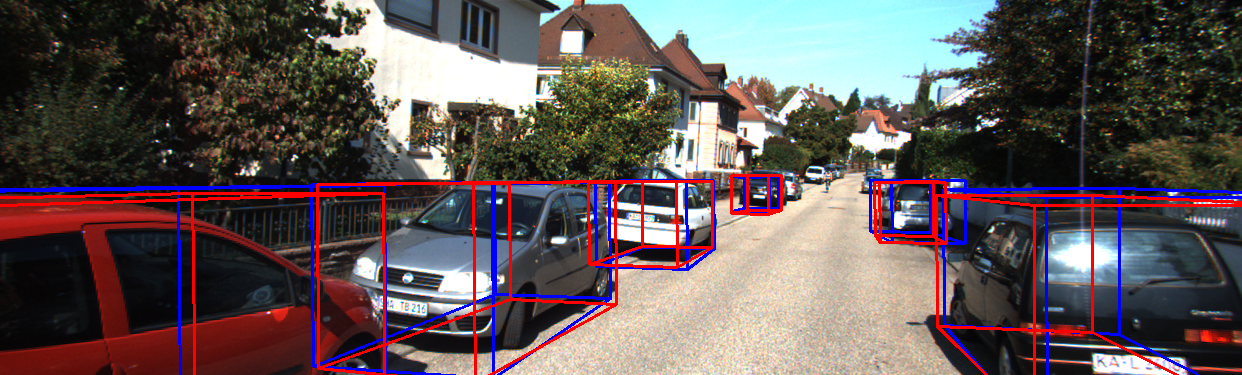}}
\end{minipage}
\hfill
\begin{minipage}{0.33\linewidth}
  \centerline{\includegraphics[width=1.0\textwidth]{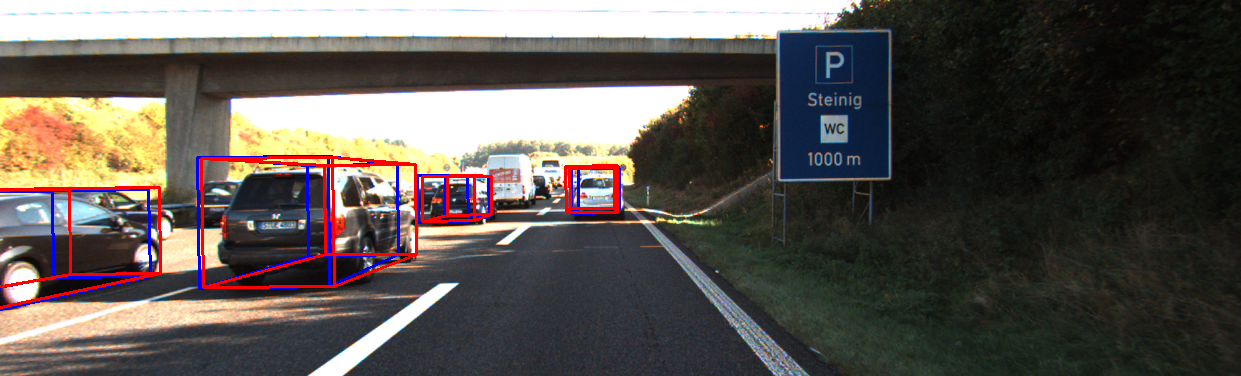}}
\end{minipage}
\hfill
\begin{minipage}{0.33\linewidth}
  \centerline{\includegraphics[width=1.0\textwidth]{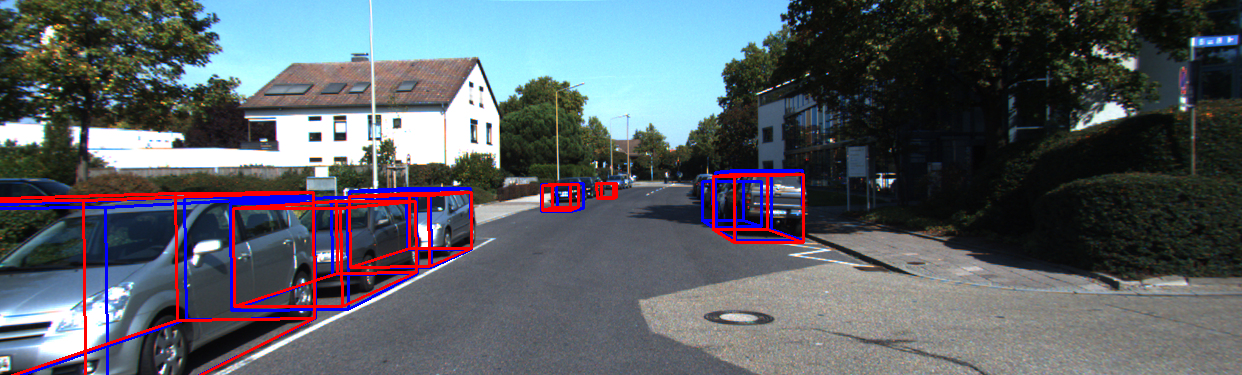}}
\end{minipage}
\vfill

\begin{minipage}{0.33\linewidth}
  \centerline{\includegraphics[width=1.0\textwidth]{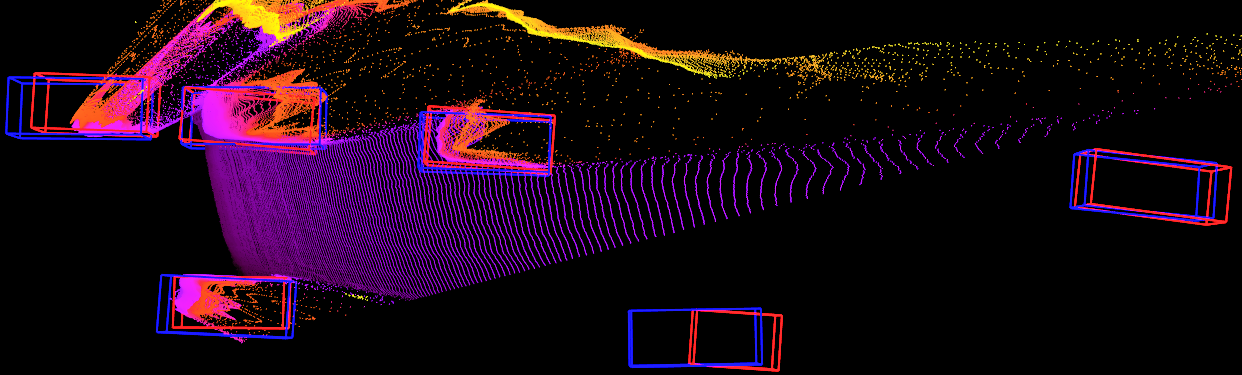}}
\end{minipage}
\hfill
\begin{minipage}{0.33\linewidth}
  \centerline{\includegraphics[width=1.0\textwidth]{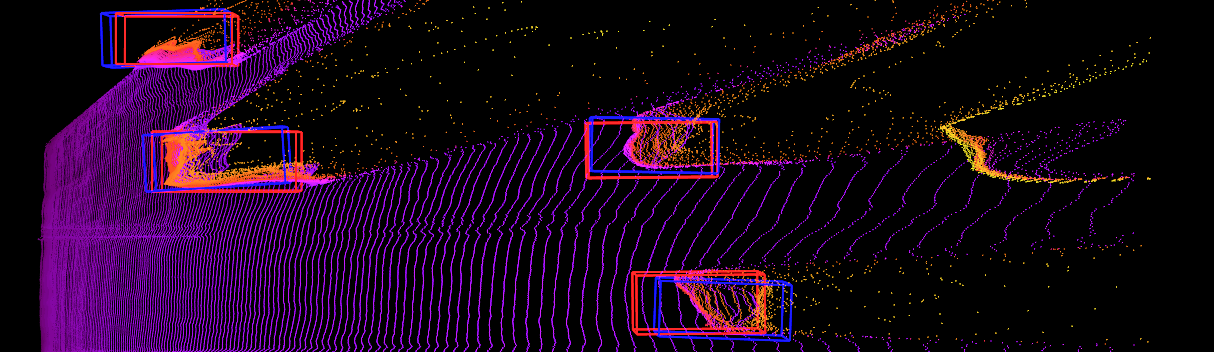}}
\end{minipage}
\hfill
\begin{minipage}{0.33\linewidth}
  \centerline{\includegraphics[width=1.0\textwidth]{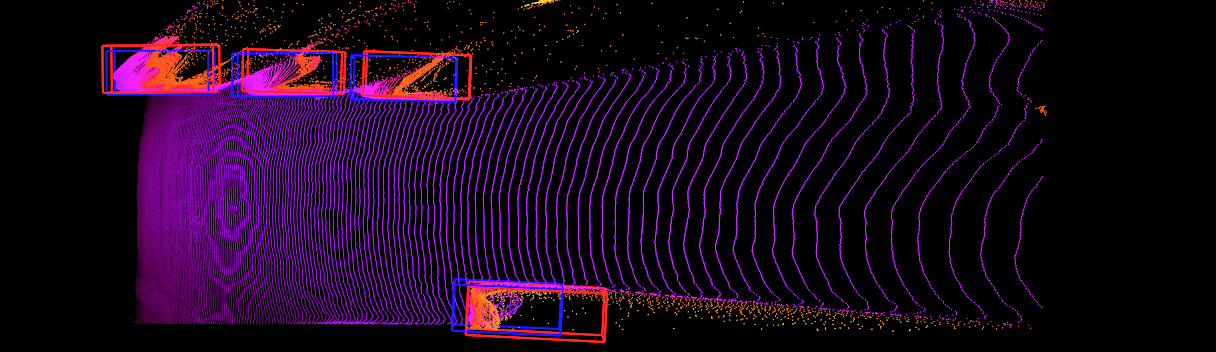}}
\end{minipage}
\caption{Qualitative results of the proposed SM3D network on KITTI val set. We visualize our 3D bounding box estimate (in blue)
and ground truth (in red) on the frontal images (1st row) and Pseudo-LiDAR point cloud (2nd row).}
\label{fig6}
\vspace{-5mm}
\end{figure*}

\noindent
\textbf{3D Detection Module.} We report the 3D detection results of car category. As shown in Table 2, we compare our results to state-of-the-art models with monocular frames and stereo pairs input from KITTI,  for the average precision (AP) of both bird's eye view (BEV) and 3D with the threshold IoU at 0.5 and 0.7, respectively. First, it is obvious that the later Pseudo-Lidar methods outperform all other "2D-3D" approaches. Inheriting such prototype, the proposed SM3D network outperforms all Pseudo-Lidar strong baselines on the average precision (AP) of both BEV and 3D for IoU threshold of 0.7, implies that SM3D performs well for challenging cases. Compared to state-of-the-art strong baseline Mono3DPLIDAR \cite{Mono3DPLiDAR}, our SM3D is 13.2\% and 11.8\% better on AP$_{BEV}$ and AP$_{3D}$, respectively. More importantly, our one-stream 3D detection module is more efficient compared to the two-stream structure of Mono3DPLIDAR\cite{Mono3DPLiDAR}, which has an additional 2D instance segmentation subbranch and network. We claim that there is no need to sacrifice the model efficiency by using "2D-3D" joint supervision training approach.  Computationally expensive strategies such as instance or semantic segmentation is not needed as the one-stream end-to-end training strategy achieves equal and better accuracy while keeping an efficient structure design. 

In our ablation study, we compare AP of end-to-end trained model (SM3D in Table 2), to non-end-to-end trained model (Naive SM3D in Table 2). For end-to-end training, SM3D is 12.7\% better on AP$_{BEV}$. For AP$_{3D}$, once being trained end-to-end, we achieve a large improvement of 27.7\%, which further validates the significance of our end-to-end training strategy. From Table 2, it is clear that large AP performance gap exists between monocular and stereo pairs input.  However, compared to other state-of-the-art monocular models, the proposed SM3D further narrows the performance gap to the state-of-the-art stereo model PL-STEREO-FP\cite{PL-MONO}. More importantly, from Table 3, our Detection Module is 2.3 times faster than PL-STEREO-FP (295 ms/frame vs 602 ms/frame).

\begin{table}[]
\caption{Inference time of each module and model.}
\setlength\tabcolsep{3pt}
\begin{tabular}{ccccc}
\hline
Module    & Mapping & Detection & SM3D & PL-STEREO\cite{PL-MONO} \\ \hline
time (ms) &82       &267         & 295          & 602  \\ \hline
\end{tabular}
\vspace{-7mm}
\end{table}

\noindent
\textbf{SM3D.} We efficiently integrate the two modules to build the proposed SM3D network. Although SM3D is a multi-task model, rather than a pure 3D detector as PL-STEREO-FP, it is still more efficient (more than 2 times faster). Considering the accuracy-efficiency trade-off, the proposed SM3D is more suitable for real-time application than stereo models. Table 3 shows the inference time of each module and the final model of SM3D.  SM3D is 18.3\% faster than a linear summation of implementing Mapping and Detection module independently, which validates our efficient model design and potential impact in real-time application.

\iffalse
% Below is an example of how to insert images. Delete the ``\vspace'' line,
% uncomment the preceding line ``\centerline...'' and replace ``imageX.ps''
% with a suitable PostScript file name.
% -------------------------------------------------------------------------
\begin{figure}[htb]

\begin{minipage}[b]{1.0\linewidth}
  \centering
  \centerline{\includegraphics[width=8.5cm]{image1}}
%  \vspace{2.0cm}
  \centerline{(a) Result 1}\medskip
\end{minipage}
%
\begin{minipage}[b]{.48\linewidth}
  \centering
  \centerline{\includegraphics[width=4.0cm]{image3}}
%  \vspace{1.5cm}
  \centerline{(b) Results 3}\medskip
\end{minipage}
\hfill
\begin{minipage}[b]{0.48\linewidth}
  \centering
  \centerline{\includegraphics[width=4.0cm]{image4}}
%  \vspace{1.5cm}
  \centerline{(c) Result 4}\medskip
\end{minipage}
%
\caption{Example of placing a figure with experimental results.}
\label{fig:res}
%
\end{figure}
\fi

%\vfill
%\pagebreak

\section{Conclusion}
In this work, we present an efficient multi-task framework SM3D, taking monocular snippets as input to simultaneously estimate map of ego-motion and 3D detection/location of surrounding objects. Using extensive improvements on framework design and novel loss function, we end-to-end train both the Mapping Module and the 3D Detection Module. Extensive results of both mapping and detection show that each module of SM3D network outperforms their state-of-the-arts baselines in accuracy. More importantly, our monocular multi-task SM3D is more efficient than single-task stereo 3D detector.  The inference time is significantly faster comparing to method with separate module computation. In the future, we will further explore the inner connection between mapping and 3D connection to make the two modules help and collaborate to each other in both training and testing, so that further improve the performance in each task.

\iffalse
For future works, we have several ideas to further improve the performance. First, instead of using depthnet directly, mapping module can be jointly trained between posenet and Pseudo-Lidar generated from depthnet, which takes the advantage of 3D information. Second, detection can help mapping, dynamic objects like vehicles violate the "rigid scene assumption" of SFM training, using detection can effectively eliminate vehicles; Finally, mapping can help detection, using the precise ego-mapping, we can project the Pseudo-Lidar point cloud from the past frames to the current frame, so that benefits 3D detection with the augmented point cloud.
\fi

\clearpage
\bibliographystyle{IEEEbib}
\bibliography{draft}

\end{document}